# Summarizing Radiology Reports' Findings into Impressions

**Raul Salles de Padua**

Quod Analytics

**Imran Qureshi**

University of Texas Austin

## Abstract

Patient hand-off and triage are two fundamental problems in health care. Often doctors must painstakingly summarize complex findings to efficiently communicate with specialists and quickly make decisions on which patients have the most urgent cases. In pursuit of these challenges, we present (1) a model with state- of-art radiology report summarization performance using (2) a novel method for augmenting medical data, and (3) an analysis of our model's limitations and radiology knowledge gain. We also provide a data processing pipeline for future models developed on the the MIMIC CXR dataset. Our best performing model was a fine-tuned BERT-to-BERT encoder-decoder with 58.75/100 ROUGE-L F1, which outperformed specialised checkpoints with more sophisticated attention mechanisms. We investigate these aspects below.

## 1 Introduction

Text summarization helps people devote attention to the most important parts of books, large bodies of text and documents. In radiology reporting, not only doctors must painstakingly summarize complex findings to efficiently communicate with specialists, but patients face a great burden of reading through a long and technical report which in most cases they don't have a full understanding of its content. Thus, the task of reports summarization is extremely relevant to radiology reports in addition to a shortage in the number of practicing radiologists in the United States[1].

Despite the vital importance of medical summarization, advances in NLP are rarely applied to this task, particularly in radiology. As a result there is very little known about language models specifically trained for radiology summarization.

In this work, we contribute solutions to the free-text radiology reports summarization task with a new state-of-art fine-tuned BERT-based model[2] in the MIMIC-CXR dataset[3]. Our model takes multiple free-text radiology report fields as input and uses a sequence-to-sequence architecture to output abstract summaries.

Our contributions are three-fold:

1. We built Biomedical-BERT2BERT, a model with state-of-art performance on the radiology text summarization task to help radiologists generate concise impressions from reports
2. A novel data augmentation strategy to improve performance on related tasks with MIMIC CXR reports
3. An analysis of the Biomedical-BERT2BERT's performance, knowledge gain, and limitations with respect to disease distribution and other architectures

We have also provided our data processing pipeline and model training code [4].

## 2 Related Work

Initial work in this domain was conducted by Chen, Gong, Zhuk [5], who reported promising results in predicting Radiologist impressions from raw findings using fine-tuned BERT-based encoder-decoder models. We extend this work by experimenting with different architectures, understanding the limitations of applying language models in this domain, and investigate the effectiveness of modern linear attention mechanisms on this dataset.

### 2.1 Tokenizers

BERT-based models have been trained on word-splits tokenizers on several corpora, mainly wiki-data and literature datasets in process usually called tokenization. Tokenization is breaking the raw text into small chunks. Tokenization breaks the raw text into words, sentences called tokens. These tokens help in understanding the context or developing the model for the NLP. The tokenization helps in interpreting the meaning of the text by analyzing the sequence of the words.

### 2.2 Pre-Trained Language Model (PLMs)

PLMs are large neural networks that are used in a wide variety of NLP tasks. They operate under a pretrain-finetune paradigm: models are first pre-trained over a large text corpus and then fine-tuned on a downstream task using additional datasets. Most common architectures, such as BERT and T5, have not been pre-trained on specialized medical corpora. We have fine-tuned our model in the MIMIC-CXR dataset, which is a large publicly available dataset of chest radiographs, free-text radiology reports, and structured labels.

### 2.3 Evaluation metrics

We evaluate summarization generation performance with Recall-Oriented Understudy for Gisting Evaluation, or ROUGE [6] on F1 metrics. Historically, ROUGE has shown good correlation with human-evaluated summaries and is a canonical metric for summarization evaluation. We focus on a variant of ROUGE, called ROUGE-L which measures the longest common subsequence overlap between the predicted and reference summaries to evaluate the informativeness of the summary.

## 3 Approach

### 3.1 Text Summarization

Our task, text summarization for biomedical documents, can be approached by either extractive or abstractive methods. Extractive summaries are snippets directly from the source text that best preserve the salient aspects of the document. Abstractive summaries, on the other hand, may generate text which may not be included in the source text.

Biomedical summaries require an abstractive approach. Radiology settings in particular require interpretation of many data points to identify the most important aspects of a patient's condition and implications for future care. For instance, a radiologists report may include only physical details of lung nodules, but the summary may conclude that "pneumonia is or is not present".

### 3.2 Baselines

We began by leveraging several pre-trained models from HuggingFace and built our own data processing pipeline[4] to extract sections and identify relevant training data from the MIMIC-CXR dataset. We baselined our project by fine-tuning a T5 encoder-decoder [7] and Facebook's BART [8] in the MIMIC-CXR dataset, implementing Teacher-Forcing[9] on our training batches with HuggingFace's DataCollatorForSeq2Seq [10]. This baseline already performed near SOTA at 47.55 ROUGE-L.

### 3.3 Main Approach

To improve upon these results, we experimented with approaches in model architecture including: 2

- Larger variants of common architectures

- Custom Encoder Decoder models
- Specialized checkpoints in medical data
- Models using linear attention mechanisms

We then moved on to a data centered approach by shuffling all input fields for each of our highest performing model architectures: T5, BERT2BERT and a BigBird-PubMed-Base model[11], with the latter model chosen because it relies on block sparse attention instead of normal attention and can handle longer sequences.

This data-centric approach proved to be key to reach a new state-of-art performance, improving the previous SOTA work[5] ROUGE-L performance by 1.38 points.

We leveraged several pre-trained models from HuggingFace and built our own data processing pipeline to extract sections and identify relevant training data from the MIMIC-CXR dataset. We have baselined our project by fine-tuning a T5 encoder-decoder [7] and Facebook' BART [8] in the MIMIC-CXR dataset.

To perform our summarization task, the input fields are fed with sentence-level embeddings created by the encoder to a Transformer decoder initialized randomly.

Encoders and Decoders are fine-tuned end-to-end, see figure 1 for this whole process.

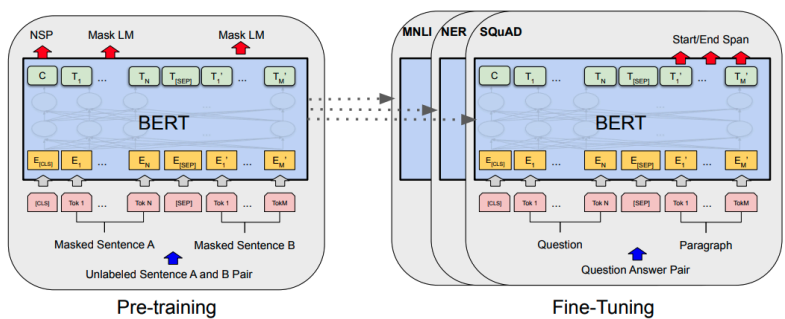

Figure 1: Pre-training and fine-tuning procedures of BERT. The same pre-trained model parameters are used to initialize models for different downstream tasks.

We use cross-entropy loss[12] with weighted average of distances to a reference word in the vector space. $Y_i$ is the i-th word in the predicted summary and $V_k$ is he k-th word in the reference vocabulary. $E(w)$ is a vector of word $w$ and $ed$ is the euclidean distance computation.

$$loss = \sum_{i=0}^{I} \sum_{k=0}^{K} p(y_{<i}, X) ed(E(V_k), E(y_i)) \qquad (1)$$

See Table 2 for an overview of model performance.

## 4  Experiments

### 4.1 Data

Our dataset is the MIMIC-CXR dataset, a collection of 377,110 chest X-ray images and 227,827 associated free-text radiology reports and structured labels[3]. The dataset is intended to support a wide body of research in medicine including image understanding, natural language processing, and decision support.

We focus on the free-text reports, each comprised of sections such as the radiologists' image obser- vations, history, comparisons between other images, final impressions, etc. For the purposes of our baselines, we sampled approx. 97,000 reports that included both FINDINGS and IMPRESSION sections, and pre-processed them to extract sections and prepare models to generate an IMPRESSION from FINDINGS.

To improve upon these results, we expanded our input fields to include FINDINGS, INDICATION, TECHNIQUE, COMPARISON. An example layout is illustrated in Figure 2. In the final phase of the project we designed a text augmentation technique by training different epochs with input fields shuffled ordering.

---

**Input Data**

COMPARISON:Chest radiograph ___
FINDINGS: Scoliosis of the thoracic spine and consequent asymmetry in the rib spaces. The compression fracture in the thoracic vertebral body is stable. Normal size of the cardiac silhouette. Normal hilar and mediastinal structures, no pulmonary edema. No pleural effusions. No pneumonia. IMPRESSION: Chronic scoliosis and stable compression fracture of a thoracic vertebra. Otherwise normal chest radiograph. No evidence pneumonia.
INDICATION: ___ year old man with two weeks of productive cough, diffuse expiratory low pitched lung sounds on exam. // r/o pneumonia,
TECHNIQUE: Chest PA and lateral

**Labels**

IMPRESSION: No evidence of pneumonia.

---

Figure 2: Example MIMIC CXR radiologist report with all input fields. Blank fields represent censored patient demographic information

### 4.2 Evaluation method

We focus on a variant of ROUGE, called ROUGE-L which measures the longest common subsequence overlap between the predicted and reference summaries to evaluate the informativeness of the summary. Equations 2 and 3 compute ROUGE precision and recall, respectively, where MaxLCS is the maximum length of longest common sequence between the reference summary (R) and the candidate summary (C). $r$ and $c$ are the lengths of the reference and candidate summaries, respectively.

$$ROUGE_{precision} = \frac{Max_{LCS}(R, C)}{r} \qquad (2)$$

$$ROUGE_{recall} = \frac{Max_{LCS}(R, C)}{c} \qquad (3)$$

$$ROUGE_{F1} = \frac{2 \; x \; Precision \; x \; Recall}{Precision + Recall} \qquad (4)$$

Precision and recall scores are combined into an F1 score as seen in equation 4. We implement

ROUGE-L leveraging the base `rouge_score` calculator available from HuggingFace: huggingface.co/metrics/rouge. Our baseline models use Cross Entropy Loss on Teacher- Forcing masked spans while training. We implemented Teacher-Forcing on our training batches with HuggingFace DataCollatorForSeq2Seq. We also used in our model BERTScore[13], to learn contextual embeddings for the reference and predicted summarization for the radiology reports and thus mitigate drawbacks of pure ROUGE score evaluation.

### 4.3 Experimental details

Our experiments established summarization from the Fine-tuned BERT2BERT model and reproduced medical summarization work by Chen et. al[5].

We used the following set of hyperparameters in the entire dataset with 80/ 20% train/ test split for our experiments in our model, illustrated below in Figure 3.

We also built a custom section extractor to automate processing whole reports into various components such as FINDINGS and IMPRESSION. These reports were filtered for those that included ALL INPUT FIELDS and IMPRESSION sections (Figure 2). The reports were then sampled, tokenized, and batched for training. Finally we shuffled inputs for last 2-3 epochs of every model trained.

| Hyperparamenters | BERT values | BigBird values | T5 values |
|---|---|---|---|
| Epochs | 6 | 7 | 12 |
| Batch Size | 8 | 8 | 2 |
| Learning Rate | 1.0e-5 | 1.0e-5 | 1.0e-5 |
| Gradient Accumulation Steps | 2 | 2 | 4 |
| Optimizer | AdamW | AdamW | AdamW |
| Training time per epoch | 0.65 hours | 1.55 hours | 0.63 hours |

Figure 3: Hyperparameters used for top 3 best performing models trained

### 4.4 Results

Our final set of results on ROUGE are shown below in Table 2. We found that a fine-tuned T5 model performs best among our baseline models. We achieved state-of-art ROUGE-L F1 performance with a BERT2BERT model after 6 epochs, with each epoch using a different ordering of input fields.

| # | Model Name | training time per epoch |
|---|---|---|
| 1 | Base - T5-Small (3 epochs) | 0 hours |
| 2 | Fine-tuned T5-Small (3 epochs) | 1.5 hours |
| 3 | DistillBART (3 epochs) | 2.5 hours |
| 4 | Base-T5-long (12 epochs) | **0.63 hours** |
| 5 | Fine-tuned BERT2BERT (6 epochs) | 0.65 hours |
| 6 | Fine-tuned PubMed BigBird (7 epochs) | 1.55 hours |
| 7 | ClinicalLongFormer2ClinicalLongFormer (4 epochs) | 1.6 hours |
| 8 | ClinicalBioBert2Transformer (previous SOTA) | - |

Table 1: Training time among different models experimented

| # | ROUGE-1 | ROUGE-2 | ROUGE-L | Baseline/ rank |
|---|---|---|---|---|
| 1 | 0.67 | 0.15 | 0.63 | Yes |
| 2 | 48.71 | 37.98 | 47.42 | Yes |
| 3 | 30.81 | 19.51 | 26.88 | Yes |
| 4 | 55.68 | **45.52** | 54.74 | 3[rd] |
| 5 | **59.61** | **48.22** | **58.75** | 1[st] |
| 6 | 57.83 | 47.12 | 56.66 | 2[nd] |
| 7 | 43.8 | 30.98 | 41.7 | - |
| 8 | 58.97 | 47.06 | 57.37 | - |

Table 2: ROUGE scores among different models we experimented with. We found that a fine-tuned BERT2BERT model performs best.

## 5    Analysis and Discussion

Our experiments yielded several non-intuitive results across model architecture, pre-training, and attention context. For instance, larger specialized models like ClinicalLongFormer [14] significantly underperformed baselines.

We investigate these results across disease types and analyze why the vanilla BERT-to-BERT model trained directly on this task outperformed models that had specialized checkpoints on clinical data, and used architectures with more sophisticated attention mechanisms.

**5.1 Summarization across Disease types**

Alongside patient radiology reports, the MIMIC-CXR dataset provides extracted disease metadata that is either indicated or negated by the radiologist. For instance, the radiologist might note that the patients chest "had no indications for pneumonia", which would be provided as "pneumonia: -1".

Comparing performance across these disease profiles (see figure 4), we observe the Biomedical- BERT2BERT model performance has a slight positive correlation with the number of examples per disease class. This indicates that while the model gains knowledge with more examples, there is potentially a model saturation point. One interpretation is that BERT-to-BERT has reached an architectural limit to improve summarization on these complex disease types.

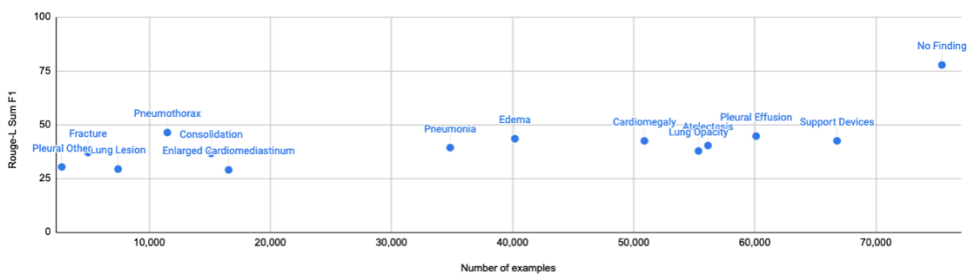

Figure 4: Performance distribution of ROUGE-L SUM scores vs. number of examples in the dataset

In contrast, the model performs almost twice as well on reports with "No Findings" (77.6 ROUGE- LSUM). "No Findings" reports are those where the radiologist still describes X-ray features, but interprets them to be normal and without disease. This summarization improvement is likely due to:

- "No Findings" reports having the largest number of examples
- Each report with no findings have a much smaller space of possible impressions compared to other disease type impressions. Other disease have a variety of nuances that are harder for the model to capture.

In fact, just summarizing reports with the phrase "There is no cardiopulmonary process" achieves a ROUGE-L of 53.2 on these reports. It is possible the model has learned an efficient classification strategy to detect No Findings reports and respond with a few canonical phrases in those cases.

**5.2 Specialized and General Checkpoints**

Interestingly, finetuned checkpoints on PubMed and other clinical data underperformed the base BERT2BERT model on this task. One example is the Clinical LongFormer model, which is pre- trained from large-scale clinical corpora [14] and achieves SOTA performance on many biomedical tasks. Similar performance was observed with BioClinical BERT [15].

However, radiology text summarization is a highly specialized task. One interpretation of this result is that other clinical checkpoints may only contain a fraction of the information required to summarize radiology notes effectively. As a result, these specialized checkpoints can easily fall into local minima with respect to the loss function, whereas a more general language checkpoint can optimize more for global minima.

For future studies, this evidence points to the importance of using a variety of pre-trained checkpoints, and not solely relying on finetuned variants for specialized tasks.

### 5.3 Limitations of Linear Attention Mechanisms

Attention is the key mechanism underlying transformers. However, the time and memory complexity to calculate attention is scales with $O(n^2)$, which restricts models like BERT to a limited context size (i.e. 512 tokens).

Many models like Linformer, Reformer, Perceiver, etc. [16] have been formulated to use linear attention methods by indirectly calculating "full attention" by approximation. Google's BigBird is the latest of such models [11] which uses random attention, windowed attention, and global attention to generate a sparse attention representation (see figure 5). The value of this approach is the ability to process 4096 tokens with sparse attention at approximately the same time complexity as with 512 tokens with full attention. Theoretically, this provides better information capture for longer documents. This is relevant for our task, as radiology reports can exceed the 512-token limit.

The BigBird authors indicate, however, that complete parity with full attention with $n$-tokens is only realized with $n$ hidden attention layers [11]. This means at $m < n$ layers, BigBird performance relies on the larger context size to have much more relevant information for the task than the 512 token limit. At $m = n$ layers, we lose the performance advantage of linear attention as $O(n \leftarrow m) = O(n^2)$.

Evaluating the information distribution in radiology text data, we find that the majority of IMPRES- SION information can be derived from only two to three sections (i.e. FINDINGS, COMPARISON, and INDICATION) whose size totalled 200-300 tokens, well within the BERT full attention limit. As a result, while BigBird might eventually achieve the Biomedical-BERT2BERT performance given more compute and scaling laws [17], the larger context size effectively acted as statistical noise versus providing an information advantage. In contrast, since we provided key sections to BERT directly, the Biomedical-BERT2BERT model learned summarization more efficiently with full attention.

For future studies, the limited effectiveness of linear attention points to the importance of evaluating the information distribution within a dataset. Likely the more concentrated relevant information is in a dataset, the less likely a larger context transformer will outperform.

### 5.4 Learning Radiology from Summarization

While transformers tend to find uninterpretable statistical patterns in the training data, we find our model has learned a few radiology facts. A few notable observations that hint at some of the operating mechanisms for Biomedical-BERT2BERT:

- Pneumonia corresponds to pleural surfaces
- Negation for disease is entailed by phrasing Normal physiology (e.g. No pneumonia = Normal heart and lungs)
- "Chest" pertains to both heart and lung anatomical features

See figure 7 for more. Visualizations are created by extracting cross-Attention matrices between our BERT2BERT Encoder Decoder components, and plotted with BERTViz [18].

We also sampled model outputs with a medical resident who found the generated summaries to encapsulate the source text well for a medical setting. See Figure 6 for model outputs examples.

This points to an exciting future direction to extract knowledge from radiology language models and provide interpretable information for users.

## 6 Conclusion

In this presented work, we built a biomedical BERT2BERT text summarization model by performing fine-tuning with also a data centricity fashion with an end-to-end deep learning approach. The model has as inputs COMPARISON, FINDINGS, IMPRESSION, INDICATION and TECHNIQUE fields and outputs IMPRESSION predictions via abstractive summarization. We believe it will help reducing human labor resources in the medical space. We used as evaluation metric ROUGE scores, to capture exact word-matching in such a sensitive patients interpretation task. We believe slight changes in this exact word matching with reference summaries may mislead patients and professionals in the medical area. Our model generates state-of-art abstractive summarization by achieving a ROUGE-L score of 58.75/ 100.

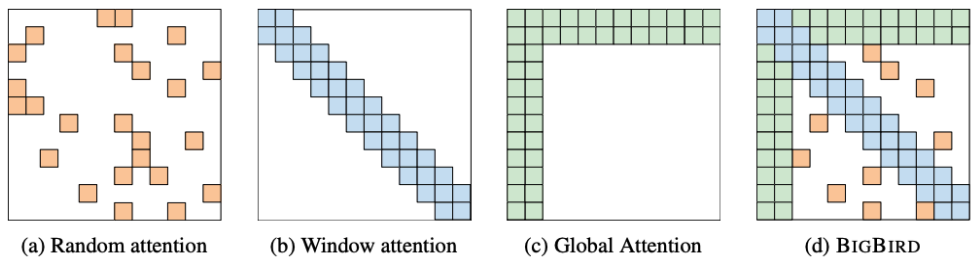

Figure 5: Multiple attention mechanisms in the BigBird linear attention calculation, which did not show improve performance for our summarization task

After thorough experimentation, we showcase that a data-centric approach signifficantly improves the quality of this task in radiology reports. Our fine-tuned model may serve as good checkpoints for other NLP endeavors in the medical space dealing with exams reports, such as summary predictions or even auto labeling.

Future work possibilities include leveraging our data centric approach and keep up addressing more data-centric opportunities for performance improvement such as treating class imbalance leaned to "No Finding" impressions. In addition, incorporating human evaluation, more specifically from radiologists, may provide deeper insights into the quality of the summaries our project and model predict. One other avenue for the future is to conduct simple baseline measurements to get a sense of the model will learn to then conduct a deep dive into thoroughly understanding the model's knowledge.

## 7    Acknoledgements

We'd like to thank all Stanford University Computer Science Department for offering all the related conceptual guidance and struture in the cloud with computing resources as well as all the staff of CS224N – Natural Language Processing with Deep Learning course.

## A Model generated outputs

| |
|---|
| Input Data |
| |
| FINDINGS: There is mild cardiomegaly. Pulmonary markings are likely accentuated by lower lung volumes. There is no consolidation or pleural effusion. No pneumothorax. There are bilateral healed rib fractures and left clavicular healed rib fracture. |
| |
| Ground Truth |
| |
| IMPRESSION: No evidence of pneumonia. |
| |
| Model Generated |
| |
| IMPRESSION: Mild cardiomegaly. No evidence of pneumonia |

Figure 6: Example radiologist report with model outputs

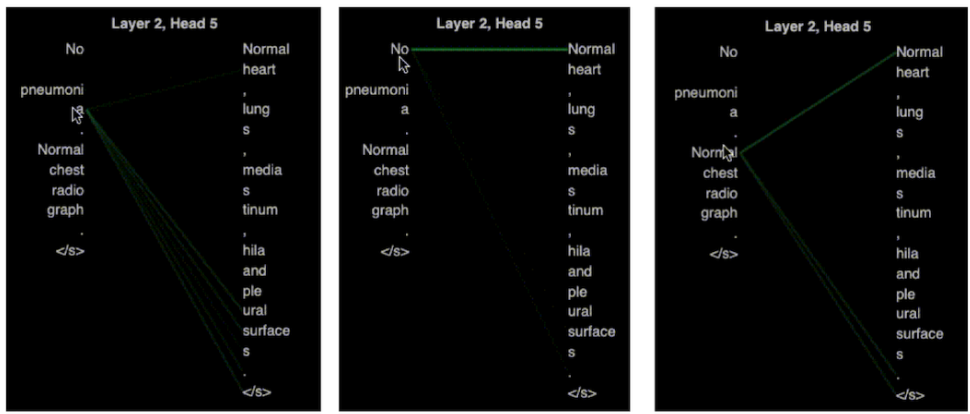

Figure 7: Visualization of BERT2BERT cross-attention weights using BERTViz [18]